\documentclass[12pt,twocolumn]{article}

\usepackage[margin=1in]{geometry}
\usepackage{graphicx}
\usepackage{caption}
\usepackage{tikz}
\usetikzlibrary{positioning,fit,arrows.meta}
\usetikzlibrary{positioning,fit,arrows.meta,calc}
\usetikzlibrary{positioning,fit,arrows.meta,calc,shapes.geometric}
\usepackage{hyperref}

\usepackage{amsmath, amssymb, amsfonts}

\usepackage{graphicx}
\usepackage{booktabs}
\usepackage{caption}
\usepackage{subcaption}

\usepackage{natbib}

\usepackage{xcolor}
\usepackage{soul}
\sethlcolor{yellow}

\usepackage{microtype}

\title{\vspace{-1em}
Probabilistic Digital Twins of Users:\\
Latent Representation Learning with Statistically Validated Semantics
\vspace{-0.5em}
}

\author{
Daniel David$^{1}$ \\
{\small $^{1}$Department of Applied Mathematics, Columbia University}
}
\date{}

\begin{document}
\maketitle

% ====================
% Abstract
% ====================
\begin{abstract}
Understanding user identity and behavior is central to applications such as personalization, recommendation, and decision support. Recent approaches typically rely on deterministic embeddings or black-box predictive models, which provide limited uncertainty quantification and little insight into what latent representations capture. In this work, we propose a probabilistic digital twin framework in which each user is modeled as a latent stochastic state that generates observed behavioral data. The digital twin is learned using amortized variational inference, enabling scalable posterior estimation while preserving a fully probabilistic interpretation.

We instantiate this framework using a variational autoencoder (VAE) applied to a rich user-response dataset designed to capture stable aspects of user identity. The model defines a generative process over observed embeddings conditioned on latent user states and learns an approximate posterior via neural variational distributions. Beyond standard reconstruction and likelihood-based evaluation, we develop a statistically grounded interpretation pipeline that links latent dimensions to observable behavioral patterns. By analyzing users at the extremes of each latent dimension and validating differences with nonparametric hypothesis tests and effect sizes, we show that specific dimensions correspond to interpretable psychological traits such as opinion strength and decisiveness.

Empirically, we find that user structure is predominantly continuous rather than discretely clustered, with weak but meaningful group structure emerging along a small number of dominant latent axes. We interpret this as a property of the underlying data-generating process and discuss how richer, intervention-driven data collection could sharpen digital twin structure. Overall, our results demonstrate that probabilistic digital twins learned via variational inference can be both expressive and interpretable, providing uncertainty-aware representations that go beyond deterministic user embeddings.
\end{abstract}

% ====================
% 1. Introduction
% ====================
\section{Introduction}

Digital twin models aim to represent individuals as latent states that summarize stable characteristics while supporting prediction, interpretation, and uncertainty quantification. In domains such as marketing, behavioral science, and decision modeling, such representations are particularly valuable when user data is high-dimensional, noisy, and heterogeneous. Recent work has emphasized learning user representations from text or response data using deterministic embeddings or deep predictive models, but these approaches often lack principled uncertainty estimates and provide limited insight into what the learned representations capture.

Probabilistic latent variable models offer a natural alternative. By defining a generative process over observations conditioned on latent user states, these models support posterior uncertainty, principled evaluation, and interpretability \citep{bishop2006pattern}. Variational autoencoders (VAEs) provide a scalable framework for learning such models via amortized variational inference \citep{kingma2013autoencoding}. While VAEs are widely used for representation learning, interpreting their latent dimensions remains challenging, and many applications rely on qualitative inspection or downstream performance alone.

In this project, we frame user modeling through the lens of \emph{probabilistic digital twins}. Each user is associated with a latent state that generates observed behavior and admits uncertainty. We treat each user as a probabilistic digital twin: a latent state that generates observed responses and admits posterior uncertainty, consistent with latent variable models of individuals in probabilistic modeling. We apply this framework to a dataset designed to capture rich, repeated user responses, closely aligned with digital twin approaches used in decision and marketing contexts. Rather than emphasizing prediction alone, we focus on understanding what the learned latent space represents.

Our primary contribution is twofold. First, we introduce a probabilistic digital twin framework that uncovers latent dimensions of user behavior through amortized variational inference. Rather than assuming predefined user types, the model learns a continuous latent space that captures underlying behavioral structure. Second, we propose a statistically validated interpretation pipeline that links individual latent dimensions to observable response patterns, enabling semantic characterization of the learned latent space. We show that while the representation exhibits weak clustering, it captures meaningful continuous variation across users. We argue that this behavior reflects the structure of the observed data and discuss how richer or more targeted measurements could improve digital twin identifiability.

% ====================
% 2. Related Work
% ====================
\section{Related Work}

\subsection{Latent Variable Models and VAEs}
Variational autoencoders provide a general framework for learning probabilistic latent representations using variational inference \citep{kingma2013autoencoding}. Extensions such as the $\beta$-VAE encourage structured latent spaces by adjusting the relative weight of the KL divergence term \citep{higgins2017beta}. VAEs have been applied to text and user data \citep{bowman2015generating}, but interpretability remains an open challenge. 

Recent work has highlighted fundamental limits on the identifiability and disentanglement of latent representations learned by unsupervised VAEs, emphasizing that latent dimensions are generally identifiable only up to rotation and rescaling without additional inductive biases or supervision \citep{locatello2019challenging,khemakhem2020vae}.

\subsection{User Modeling and Representation Learning}
User representations are commonly learned using matrix factorization \citep{koren2009matrix}, deterministic embeddings \citep{reimers2019sentence,devlin2019bert}, or autoencoders. While effective for prediction, these methods typically lack uncertainty quantification and generative semantics. 

More recent probabilistic approaches model users as latent stochastic states, enabling uncertainty-aware inference and principled reasoning about individual-level heterogeneity. This perspective aligns with emerging work on digital twin representations in human-centered systems, where latent variable models are used to summarize stable behavioral tendencies while accounting for observational noise.

\subsection{Interpretability and Validation}
Interpreting latent representations has received increasing attention \citep{li2016visualizing,bau2019network}. However, many approaches rely on qualitative inspection or heuristic probing of learned features. In contrast, our work provides statistically validated interpretations of latent dimensions using hypothesis testing and effect sizes \citep{mann1947test,cohen1988statistical}, enabling principled semantic alignment between learned representations and observable behavioral patterns.
% ====================
% 3. Model
% ====================
\section{Probabilistic Digital Twin Model}

We model each user as a \emph{probabilistic digital twin}: a latent stochastic state that summarizes stable aspects of identity and generates observed behavior. This formulation follows the latent variable modeling perspective common in probabilistic modeling, where individual-level heterogeneity is captured through continuous latent structure rather than discrete types \citep{blei2003latent,blei2007correlated}.

\subsection{Generative Model}

Let $u \in \{1,\dots,U\}$ index users. Each user is associated with a latent digital twin state
\[
z_u \in \mathbb{R}^K,
\]
which represents stable but unobserved characteristics of the user.

We place a standard Gaussian prior on latent states:
\begin{equation}
z_u \sim \mathcal{N}(0, I).
\end{equation}

Observed user data are represented as embeddings $x_u \in \mathbb{R}^D$, derived from aggregating repeated user responses. Conditioned on the latent state, observations are generated via a probabilistic decoder:
\begin{equation}
x_u \mid z_u \sim \mathcal{N}(f_\theta(z_u), \sigma^2 I),
\end{equation}
where $f_\theta(\cdot)$ is a neural network parameterizing the mean of the observation distribution. This defines a fully generative model in which observed behavior arises as a noisy manifestation of an underlying digital twin.

\begin{figure}[t]
\centering
\begin{tikzpicture}[
    latent/.style={ellipse, draw, thick, minimum width=1.3cm, minimum height=0.9cm},
    obs/.style={rectangle, draw, thick, rounded corners, minimum width=1.6cm, minimum height=0.9cm},
    block/.style={rectangle, draw, thick, minimum width=1.6cm, minimum height=0.9cm},
    arrow/.style={->, thick},
    plate/.style={draw, thick, rounded corners}
]

% Nodes
\node[latent] (z) {$z_u$};
\node[block, right=1.4cm of z] (f) {$f_\theta$};
\node[obs, right=1.4cm of f] (x) {$x_u$};

% Generative arrows
\draw[arrow] (z) -- (f);
\draw[arrow] (f) -- node[above, font=\scriptsize] {$p_\theta(x_u \mid z_u)$} (x);

% Prior
\node[block, above=0.6cm of z, xshift=0.4cm] (pz) {$p(z)$};
\draw[arrow] (pz) -- (z);

% Inference (dashed)
\node[block, below=0.9cm of f] (q) {$q_\phi(z_u \mid x_u)$};
\draw[arrow, dashed] (x) -- (q);
\draw[arrow, dashed] (q) -- (z);

% Plate
\node[plate, fit=(z)(f)(x)(q), inner sep=0.35cm] (userplate) {};
\node[anchor=south east, font=\scriptsize] at (userplate.north east) {$u=1,\ldots,U$};
\end{tikzpicture}

\caption{
Block and plate diagram of the probabilistic digital twin model.
For each user $u$, a latent digital twin state $z_u \sim p(z)$ is transformed by a neural decoder
$f_\theta$ to parameterize the observation distribution $p_\theta(x_u \mid z_u)$, generating
observed behavior $x_u$.
Inference is performed via an encoder $q_\phi(z_u \mid x_u)$ (dashed arrows).
}
\label{fig:block_plate}
\end{figure}
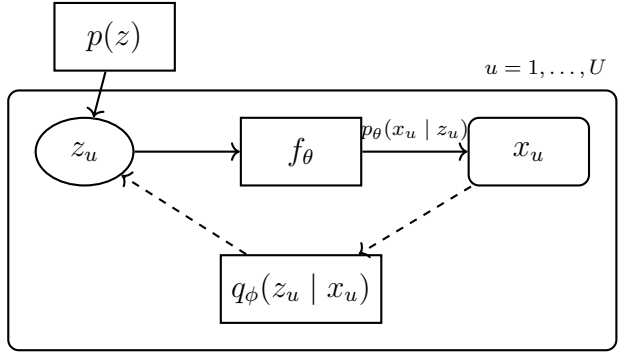

\subsection{Variational Inference}

Exact posterior inference for $p(z_u \mid x_u)$ is intractable. We therefore use amortized variational inference \citep{blei2017variational}, introducing a variational family
\begin{equation}
q_\phi(z_u \mid x_u) = \mathcal{N}\!\left(\mu_\phi(x_u), \Sigma_\phi(x_u)\right),
\end{equation}
where $\mu_\phi(\cdot)$ and $\Sigma_\phi(\cdot)$ are parameterized by a neural encoder.

Model parameters $\theta$ and variational parameters $\phi$ are learned by maximizing the evidence lower bound (ELBO):

\begin{align}
\mathcal{L}
&= \mathbb{E}_{q_\phi(z_u \mid x_u)}
   \big[\log p_\theta(x_u \mid z_u)\big] \nonumber \\
&\quad - \beta \,
   \mathrm{KL}\!\left(q_\phi(z_u \mid x_u)\,\|\,p(z_u)\right).
\label{eq:elbo}
\end{align}

The hyperparameter $\beta$ controls the strength of regularization and encourages structured latent representations \citep{higgins2017beta}. During training, we observed KL collapse in early experiments, indicating under-utilization of the latent space. We mitigate this by constraining the encoder variance and tuning $\beta$, resulting in stable training and non-degenerate posteriors.

\subsection{Interpretation as Digital Twin Inference}

Under this formulation, the encoder network does not define the model itself but instead parameterizes an approximate posterior over digital twin states. The learned latent variables therefore admit a clear probabilistic interpretation: $q_\phi(z_u \mid x_u)$ represents uncertainty over a user’s digital twin conditioned on observed behavior.

This perspective enables principled analysis of the latent space, including uncertainty-aware comparisons between users and downstream interpretability analysis.
% ====================
% 4. Dataset and Experimental Setup
% ====================
\section{Dataset and Experimental Setup}

We evaluate our model on the \emph{Twin-2K-500} user-response dataset \citep{twin2k500}, which is designed to capture stable aspects of individual preferences and attitudes through repeated measurements. For each user, observed responses are summarized as a high-dimensional embedding, which we treat as a noisy observation generated by an underlying latent digital twin. The embedding is not assumed to be a deterministic representation of the user, but rather a stochastic manifestation of their latent state, aggregated across many responses.

This dataset is well suited for probabilistic digital twin modeling, as it emphasizes consistency across a large number of measurements rather than sparse or transactional interactions. Under this interpretation, the embedding serves as input to amortized variational inference, while the latent variable represents the user’s digital twin. Additional details on dataset composition and preprocessing are provided in Appendix~A.

We compare our VAE-based digital twin model against PCA \citep{jolliffe2016principal}, factor analysis \citep{bartholomew2011latent}, and deterministic baselines. Evaluation metrics include reconstruction error, KL divergence, and qualitative analysis of latent space structure.

% ====================
% 5. Results
% ====================
\section{Results}

\subsection{Model Fit and Latent Usage}

\begin{figure}[t]
  \centering
  \includegraphics[width=\columnwidth]{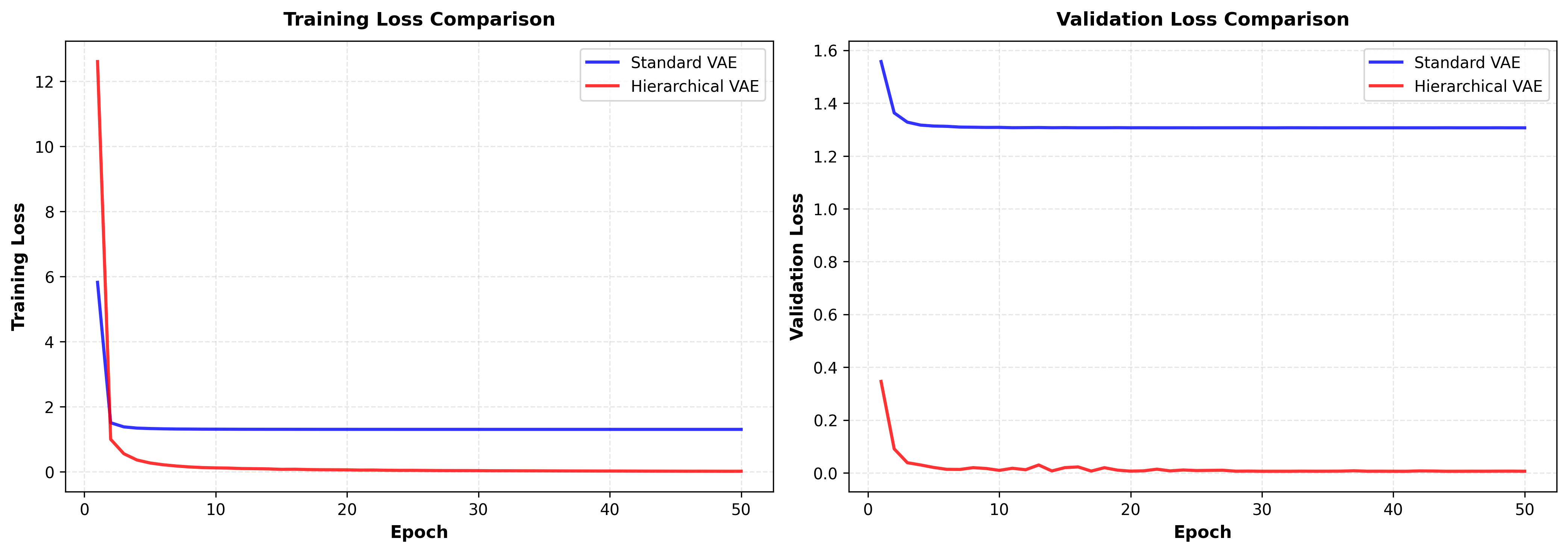}
  \caption{
  Training and validation loss curves for the standard and hierarchical VAE models.
  The hierarchical VAE converges more rapidly and achieves lower loss on both training
  and validation data, indicating improved model fit.
  }
  \label{fig:training_curves}
\end{figure}

Figure~\ref{fig:training_curves} shows training and validation loss curves for the standard and hierarchical VAE models. Both models converge rapidly, with the hierarchical VAE achieving substantially lower loss on both training and validation data, consistent with prior work on hierarchical variational models \citep{ranganath2016hierarchical}. The stable convergence behavior and consistent gap between the two models indicate improved model fit rather than overfitting.

While the loss is optimized as a single variational objective, the observed training dynamics are consistent with non-degenerate utilization of the latent space. In particular, the absence of degenerate convergence and the improved generalization performance suggest that the learned latent variables capture meaningful structure beyond observation noise.

\subsection{Latent Space Structure}

\begin{figure}[t]
  \centering
  \includegraphics[width=\columnwidth]{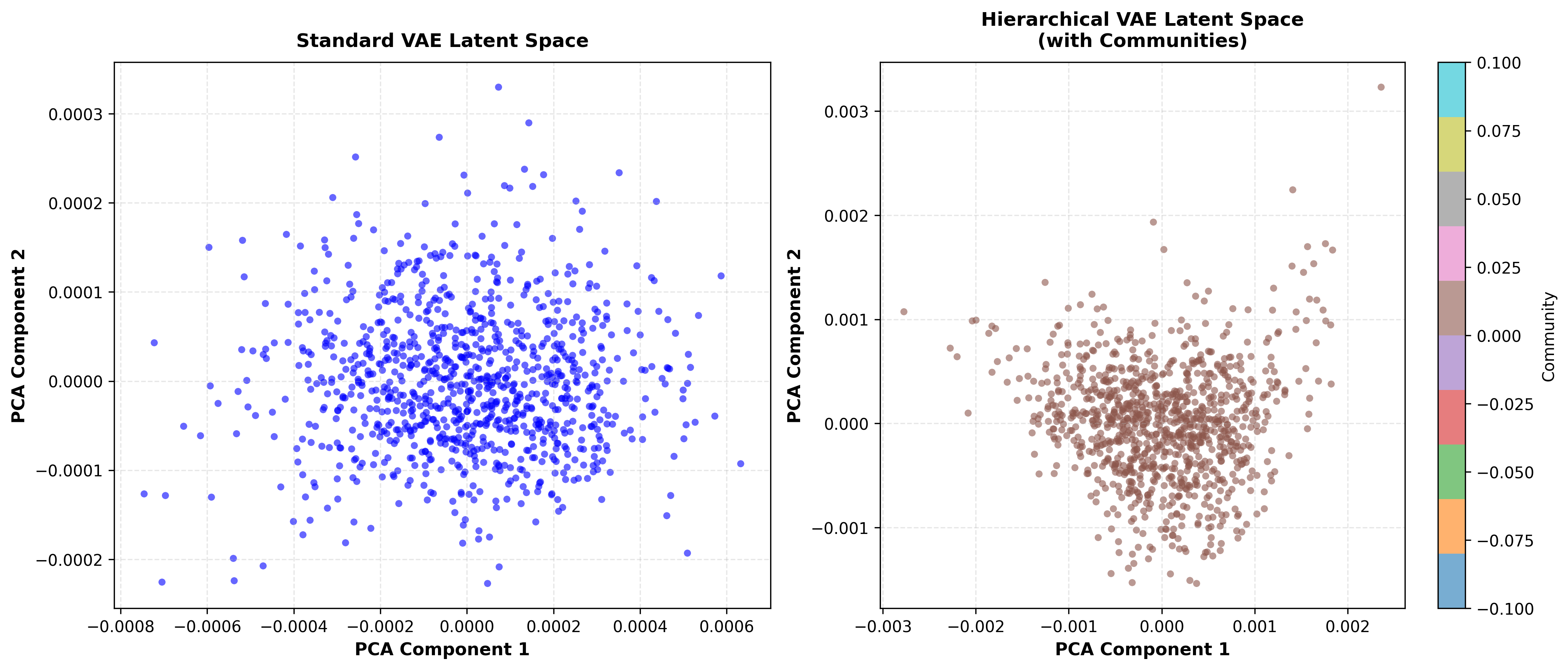}
  \caption{
  Low-dimensional visualization of the learned latent space using PCA.
  Each point corresponds to a user and is colored by the dominant latent dimension.
  The latent structure appears predominantly continuous, with variation organized along a small number of axes.
  }
  \label{fig:latent_space}
\end{figure}

Visualization of the learned latent representations in Figure~\ref{fig:latent_space} reveals a predominantly continuous latent structure with weak global clustering. Variation across users is organized along a small number of dominant latent dimensions, rather than forming sharply separated groups. Notably, this qualitative structure is consistent across both the standard and hierarchical VAE variants, whose PCA projections appear similar for this use case. This suggests that differences in user behavior are better characterized by continuous axes of variation than by discrete user types, independent of whether hierarchical structure is imposed.

These observations support the use of a continuous latent digital twin representation and are consistent with the smooth, high-dimensional structure of the observed response data. We next examine whether individual latent dimensions admit statistically validated semantic interpretations.

\subsection{Statistically Validated Interpretation}

To interpret the learned latent space, we analyze users at the extremes of each latent dimension and compare their response patterns. Dimension~33 exhibits highly significant differences between high and low groups, corresponding to variation in opinion strength and decisiveness. These differences are consistent across multiple response categories and are statistically significant with large effect sizes.

\begin{figure}[h]
  \centering
  \includegraphics[width=\columnwidth]{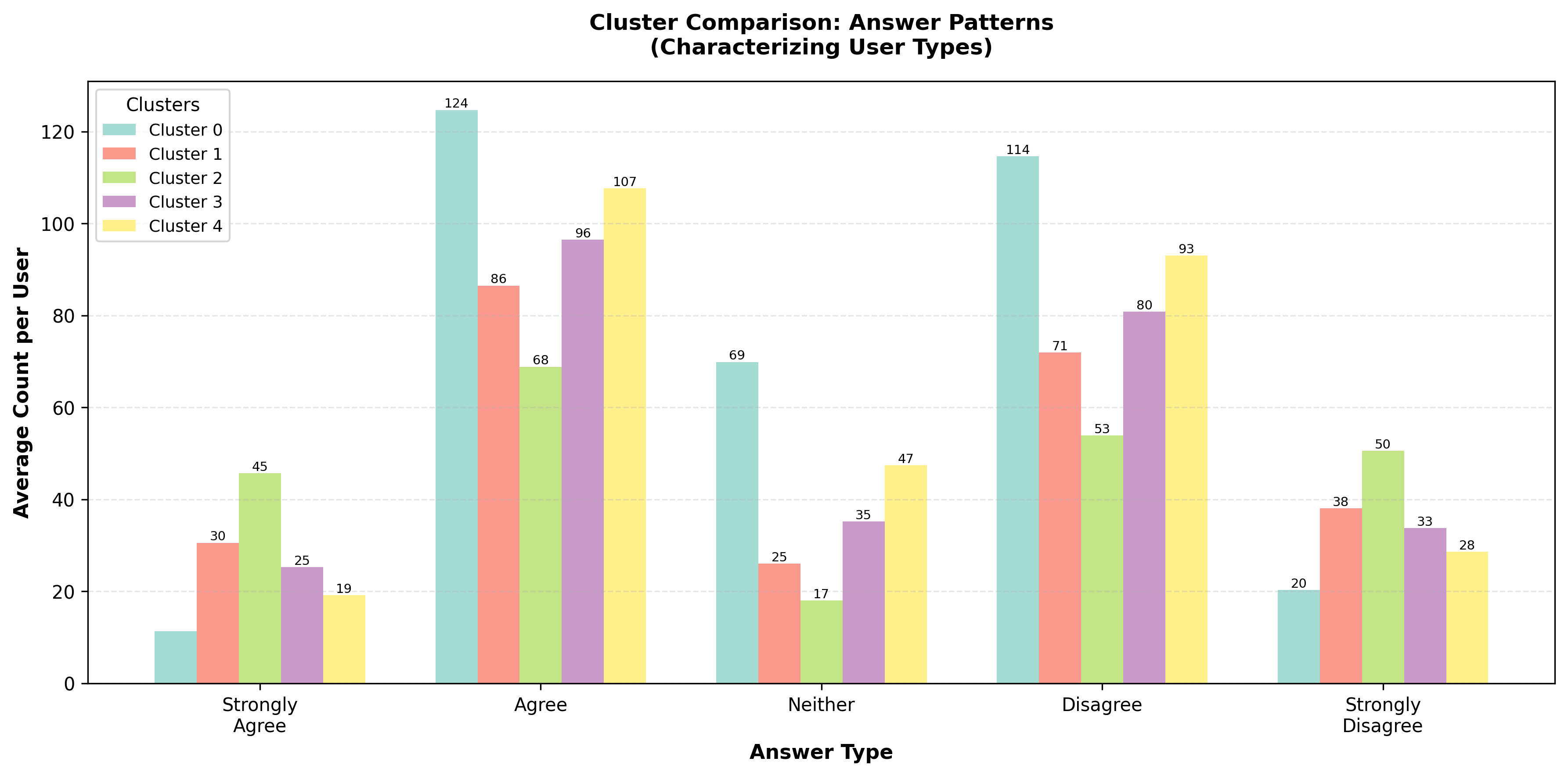}
  \caption{
  Cluster-wise comparison of response extremity versus neutrality.
  Clusters primarily reflect variation along the dominant latent dimension rather than sharply distinct user groups, consistent with a continuous latent structure.
  }
  \label{fig:cluster_extremity}
\end{figure}

\begin{figure}[t]
  \centering
  \includegraphics[width=0.9\columnwidth]{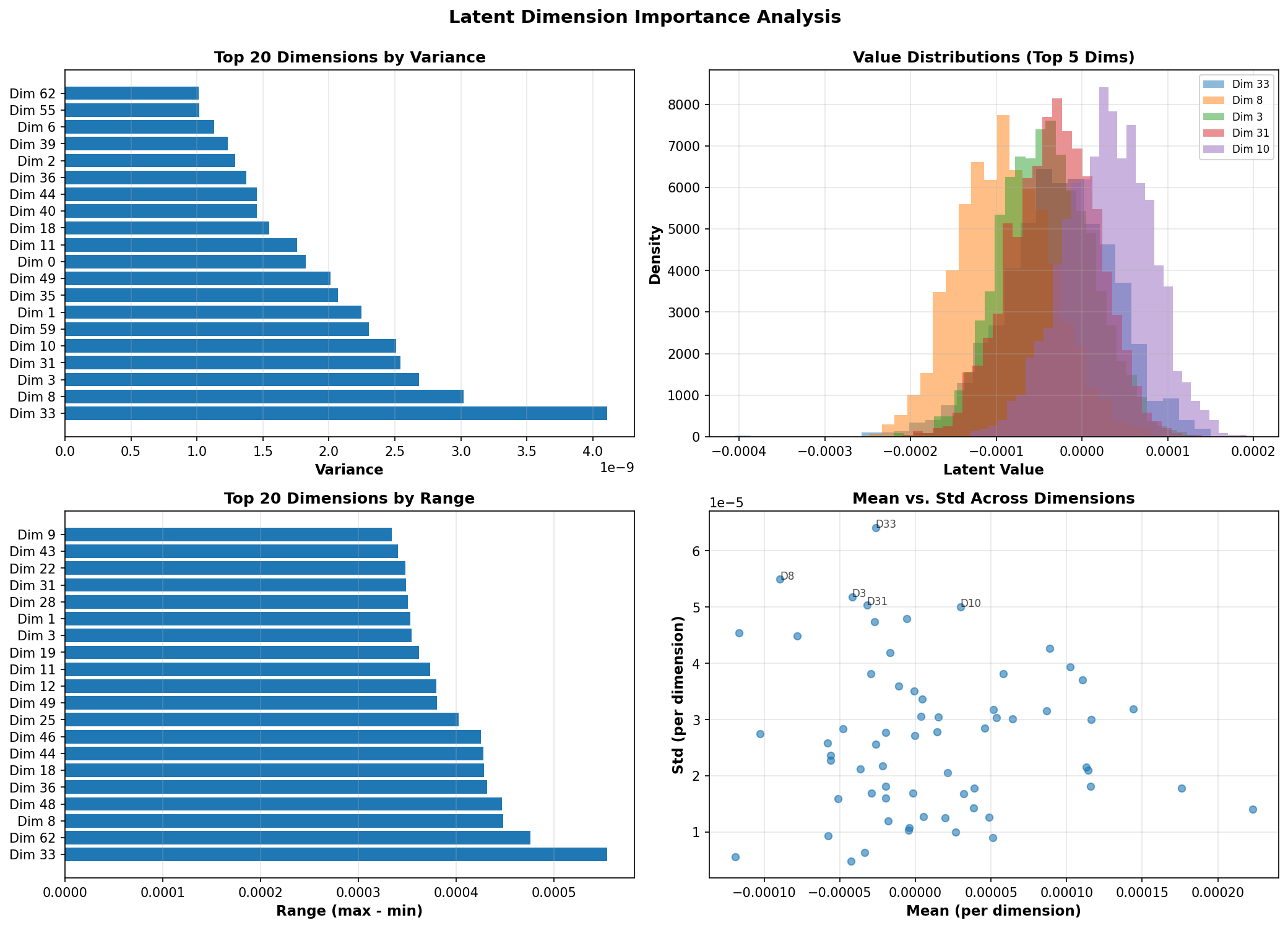}
  \caption{
  Latent dimension importance analysis.
  Shown are variance- and range-based rankings across latent dimensions, along with distributional summaries
  for a subset of high-importance dimensions.
  Dimension~33 consistently emerges as the most salient axis under multiple quantitative criteria,
  motivating its selection for detailed interpretation.
  }
  \label{fig:dimension_importance}
\end{figure}

Cluster analysis in Figure~\ref{fig:cluster_extremity} further confirms that the observed weak clusters align primarily with this dominant latent dimension. Rather than indicating the absence of structure, this pattern suggests that user variation in this dataset is organized along a small number of continuous axes with only weakly expressed group-level separation. From a digital twin perspective, this supports representations that accommodate both continuous latent variation and higher-level aggregation, with the relative prominence of each determined by the richness of the observed data. Additional evidence supporting the salience of this latent axis is provided in \hyperref[app:extended]{Appendix~B}.

We emphasize that latent Dimension~33 should not be interpreted as a predefined or intrinsically privileged feature. Rather, it represents one axis of variation learned by the VAE through data-driven feature extraction under the variational objective. Because the latent space is identifiable only up to rotation and rescaling, semantic information may be distributed across dimensions, and the prominence of Dimension~33 reflects alignment between a learned latent direction and a dominant mode of variation in the data. From a digital twin perspective, the goal is not to identify a single meaningful dimension, but to recover a latent representation in which behavioral factors are expressed along learned latent scales with quantified uncertainty. The interpretability of Dimension~33 demonstrates that such alignment is possible, but does not imply uniqueness; meaningful digital twin representations arise when latent dimensions collectively span interpretable axes of variation.

\begin{figure}[htbp]
  \centering
  \includegraphics[width=0.9\columnwidth]{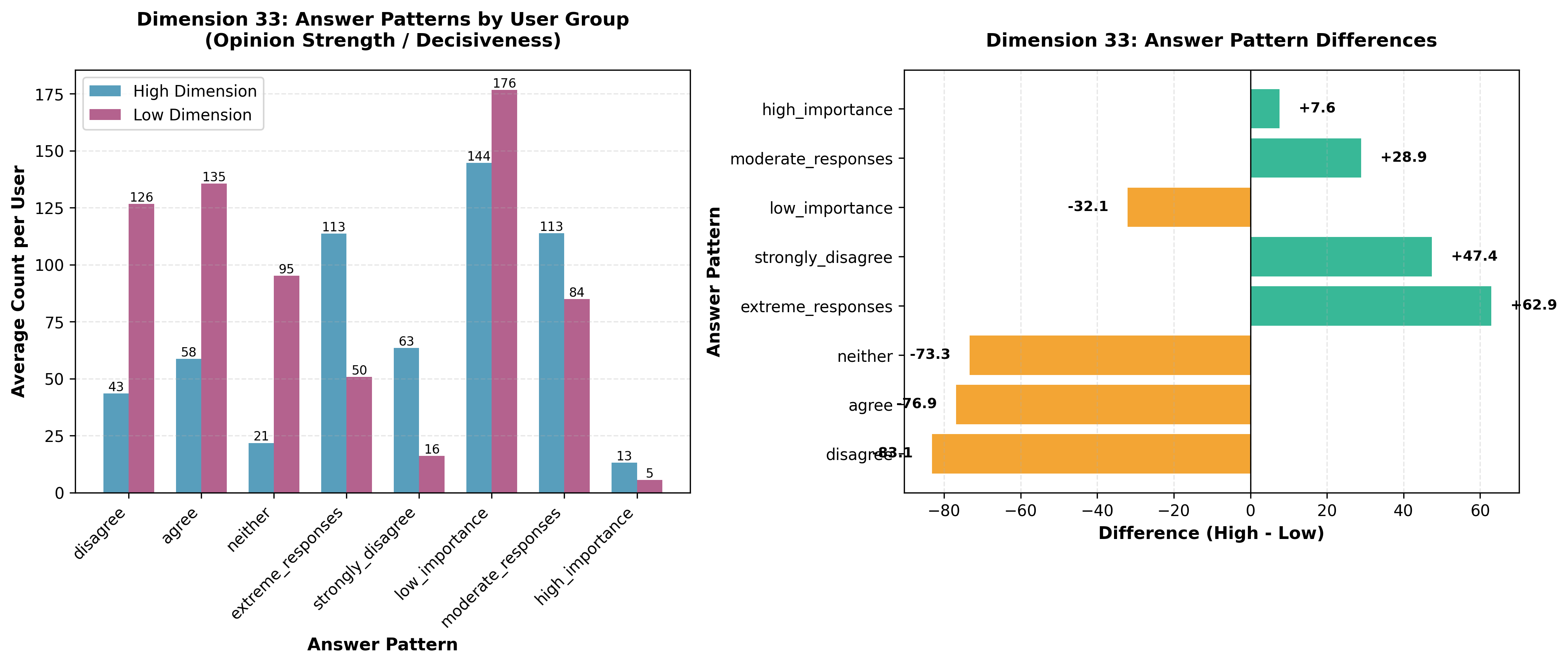}
  \caption{Statistically validated interpretation of Dimension 33. 
  Users with high latent values (top) show significantly more extreme 
  responses (e.g., ``Strongly Agree/Disagree'', +62.9 instances, $p < 0.001$) 
  and fewer neutral options (-73.3 instances of ``Neither'', $p < 0.001$), 
  while low-value users (bottom) prefer moderate responses. 
  This dimension captures \emph{opinion strength} and decisiveness, 
  with large effect size (Cohen's $d > 0.8$).}
  \label{fig:dim33}
\end{figure}

% ====================
% 6. Discussion
% ====================
\section{Discussion}

Our findings suggest that user digital twins learned from this dataset are naturally represented along continuous latent dimensions, capturing graded variation in behavior rather than sharply separated types. From a digital twin perspective, such continuity is beneficial, as it enables nuanced representation and uncertainty-aware inference. Sharper or partially discrete structure may emerge under richer data regimes, such as adaptive questioning or intervention-based measurements that elicit stronger behavioral signals.

The statistically validated interpretation pipeline demonstrates that probabilistic latent spaces can be meaningfully interpreted, addressing a common criticism of VAEs as black boxes.

We additionally explored hierarchical extensions of the variational model to assess whether community-level structure could induce sharper latent organization. While hierarchical variants improved optimization behavior and model fit, they did not fundamentally alter the continuous nature of the learned latent space. This suggests that the absence of discrete clustering reflects limitations of the observational data rather than insufficient model expressiveness, reinforcing the importance of data design in digital twin modeling. 
At a higher level, results suggest that modeling effort alone cannot substitute for experimental or intervention-based data collection when the goal is to identify discrete user types.

% ====================
% 7. Conclusion
% ====================
\section{Conclusion}

We presented a probabilistic digital twin framework for user modeling learned via amortized variational inference. By combining generative modeling with rigorous interpretation and statistical validation, we showed that latent user representations can uncover meaningful behavioral dimensions while remaining continuous and uncertainty-aware. Our results highlight the central role of data design in shaping digital twin structure and suggest that richer, more targeted measurements may be necessary to recover fully disentangled and decision-relevant user representations.

% ====================
% Bibliography
% ====================
\bibliography{references}
\bibliographystyle{plainnat}

\newpage
\appendix
\section{Dataset and Model Details}

This appendix provides additional dataset statistics and model configuration details that are omitted from the main text for brevity.

\subsection{Dataset Statistics}

Table~\ref{tab:dataset_full} summarizes key statistics of the Twin-2K-500 dataset used in our experiments. The dataset is characterized by dense observational structure, with a large number of repeated identity-related responses per user and a smaller set of update responses.

\begin{table}[h]
\centering
\caption{
Detailed dataset statistics for Twin-2K-500.
}
\label{tab:dataset_full}
\small
\renewcommand{\arraystretch}{0.95}
\begin{tabular}{l c}
\hline
\textbf{Statistic} & \textbf{Value} \\
\hline
Number of users & 2{,}058 \\
Total identity texts & 2{,}354{,}212 \\
Total update texts & 405{,}779 \\
Identity texts per user (mean) & 1{,}143.9 \\
Identity texts per user (median) & 1{,}144.0 \\
Update texts per user (mean) & 197.2 \\
Update texts per user (median) & 197.0 \\
Identity text length (mean chars) & 71 \\
Identity text length (median chars) & 37 \\
Update text length (mean chars) & 97 \\
Update text length (median chars) & 110 \\
Embedding dimension & 384 \\
\hline
\end{tabular}
\end{table}

\subsection{Model Configuration}

Table~\ref{tab:model_full} reports the full configuration of the probabilistic digital twin model used in all experiments.

\begin{table}[h]
\centering
\caption{
Model configuration details.
}
\label{tab:model_full}
\small
\renewcommand{\arraystretch}{0.95}
\begin{tabular}{l c}
\hline
\textbf{Component} & \textbf{Specification} \\
\hline
Latent dimension ($K$) & 64 \\
Prior distribution & $\mathcal{N}(0, I)$ \\
Decoder likelihood & Gaussian \\
Observation noise & Isotropic \\
Inference method & Amortized VI \\
Encoder architecture & Neural network \\
Decoder architecture & Neural network \\
Regularization & $\beta$-VAE ($\beta = 1.0$) \\
Optimizer & Adam \\
Batch size & 32 \\
Training epochs & 50 \\
\hline
\end{tabular}
\end{table}

\subsection{Embedding Distribution Diagnostics}

\begin{figure}[h!]
  \centering
  \includegraphics[width=\columnwidth]{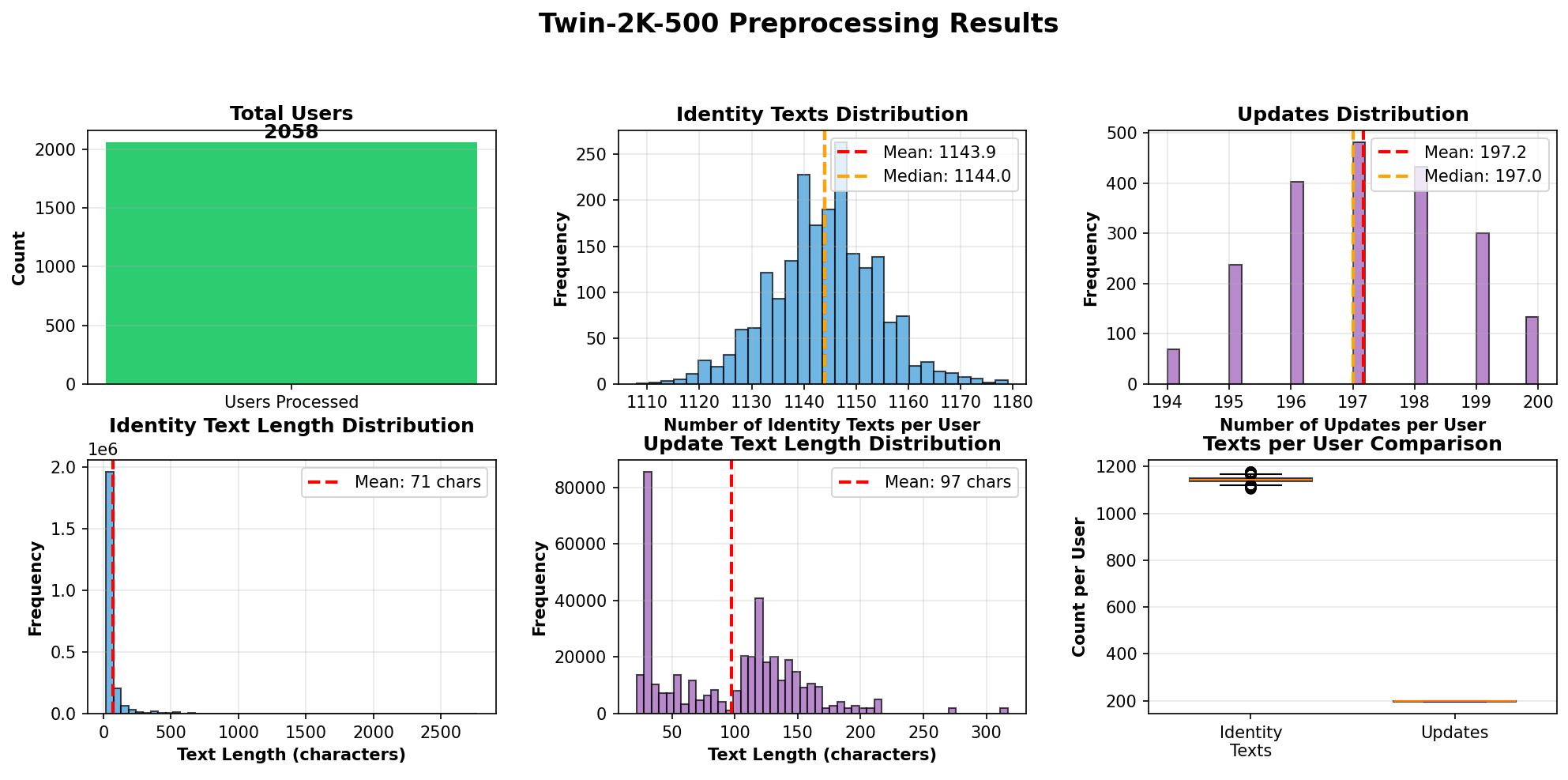}
  \caption{
 Preprocessing results for the Twin-2K-500 dataset.
The figure summarizes basic distributional properties of the processed embeddings used in subsequent modeling.}
  \label{fig:embedding_distributions_appendix}
\end{figure}

\newpage
\section{Extended Analysis and Additional Visualizations}
\label{sec:appendixB}

This appendix contains additional figures that support the results presented in the main text.
These visualizations provide further detail on latent dimension importance, response pattern differences,
and cluster-level summaries, but are not required to follow the core arguments of the paper.

\begin{figure}[t]
  \centering
  \includegraphics[width=\columnwidth]{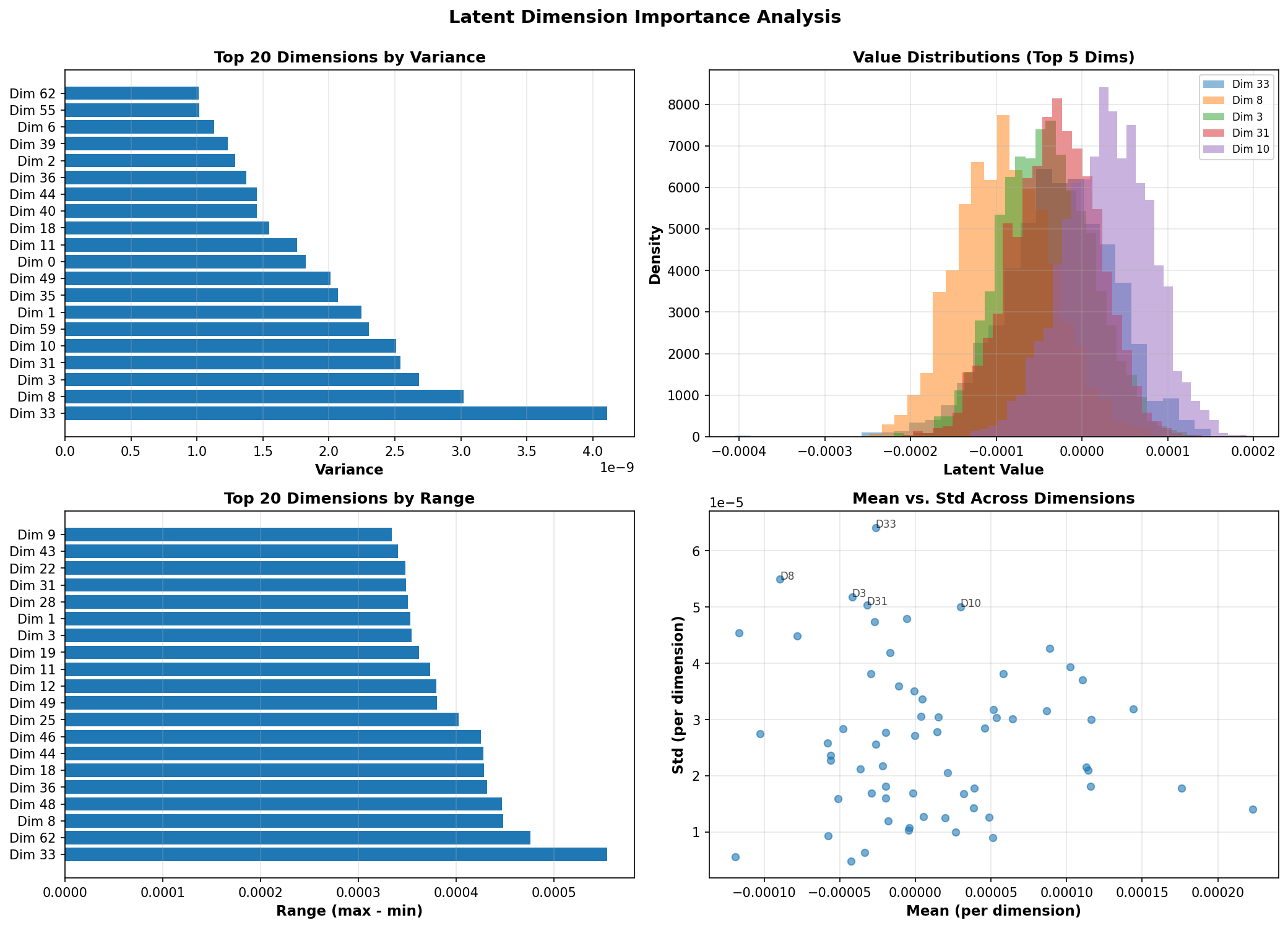}
  \caption{
  Summary statistics characterizing latent dimension importance.
  Shown are variance- and range-based rankings of latent dimensions, as well as distributional summaries
  for a subset of high-importance dimensions.
  }
  \label{fig:appendix_dimension_importance}
\end{figure}

\begin{figure}[t]
  \centering
  \includegraphics[width=\columnwidth]{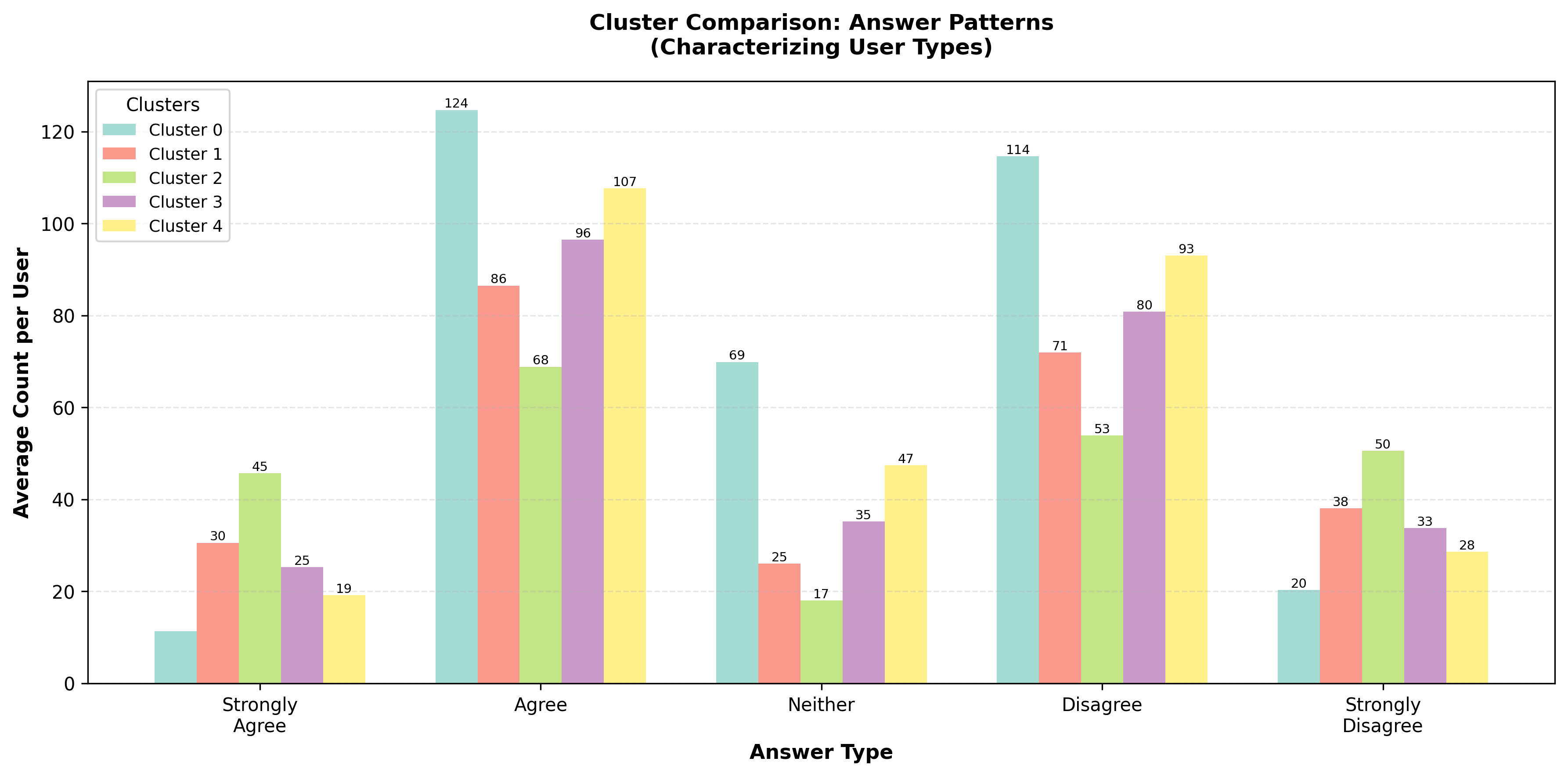}
  \caption{
  Aggregate response patterns across inferred clusters.
  Counts are averaged across users within each cluster for different response categories.
  }
  \label{fig:appendix_cluster_patterns}
\end{figure}

\begin{figure}[t]
  \centering
  \includegraphics[width=\columnwidth]{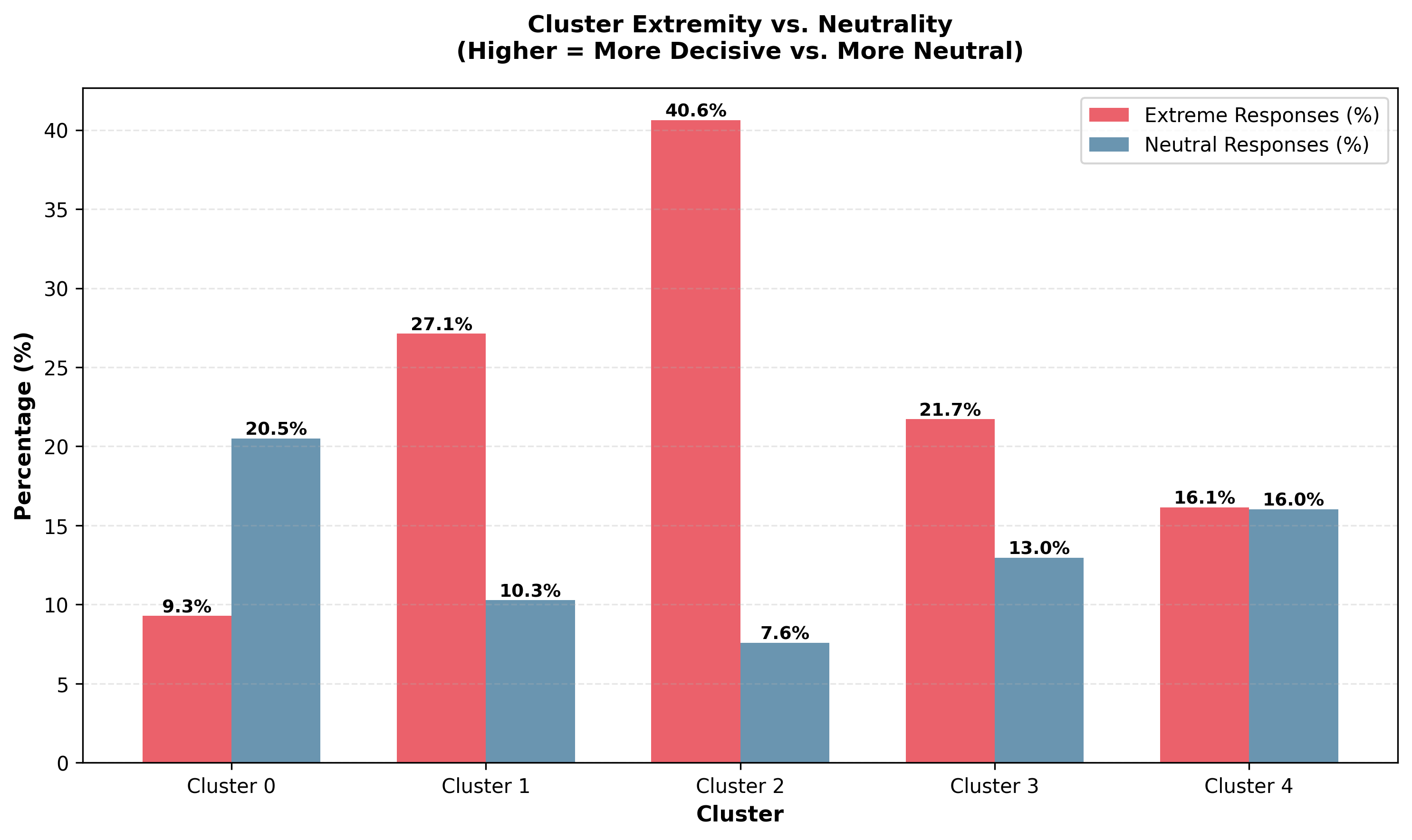}
  \caption{
  Cluster-level comparison of response extremity and neutrality.
  Bars show the relative prevalence of extreme versus neutral responses across clusters.
  }
  \label{fig:appendix_cluster_extremity}
\end{figure}

\begin{figure}[t]
  \centering
  \includegraphics[width=\columnwidth]{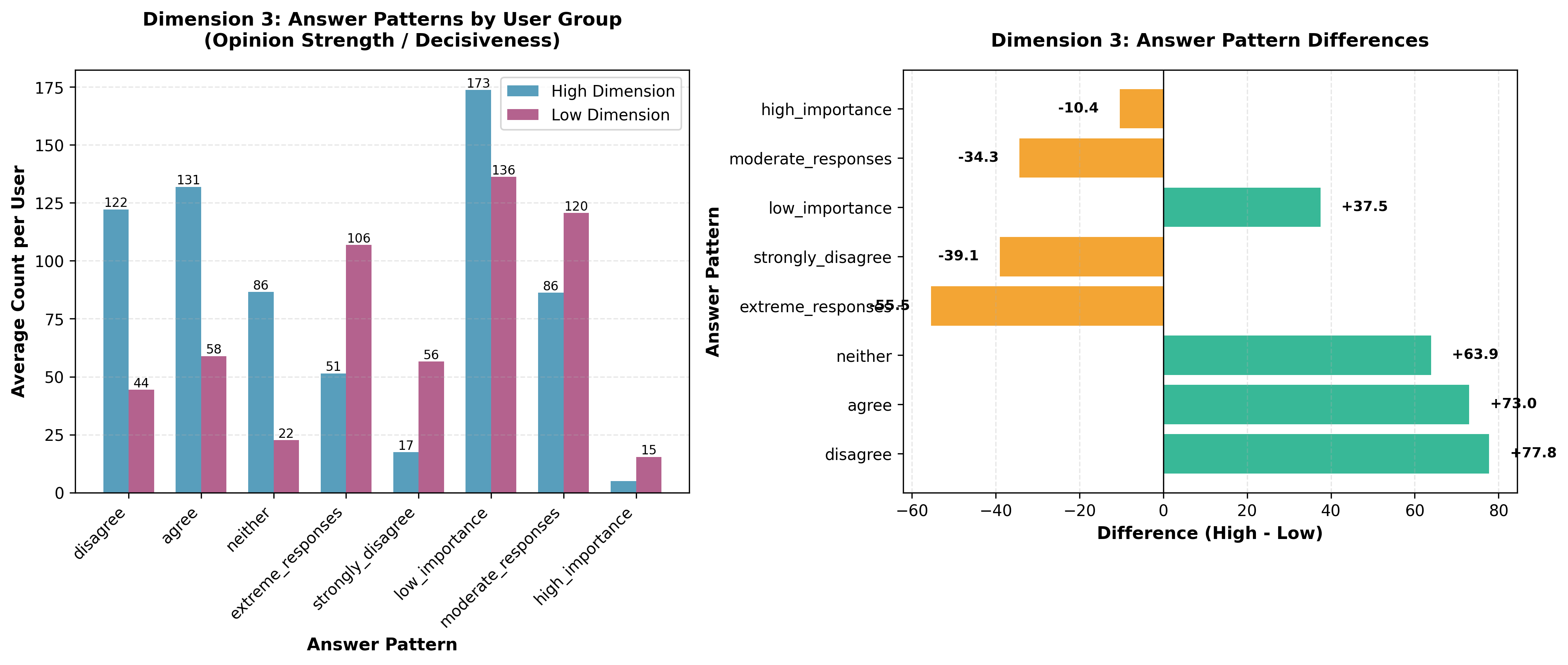}
  \caption{
  Response pattern comparison for users at the high and low extremes of latent Dimension~3.
  }
  \label{fig:appendix_dim3_patterns}
\end{figure}

\begin{figure}[t]
  \centering
  \includegraphics[width=\columnwidth]{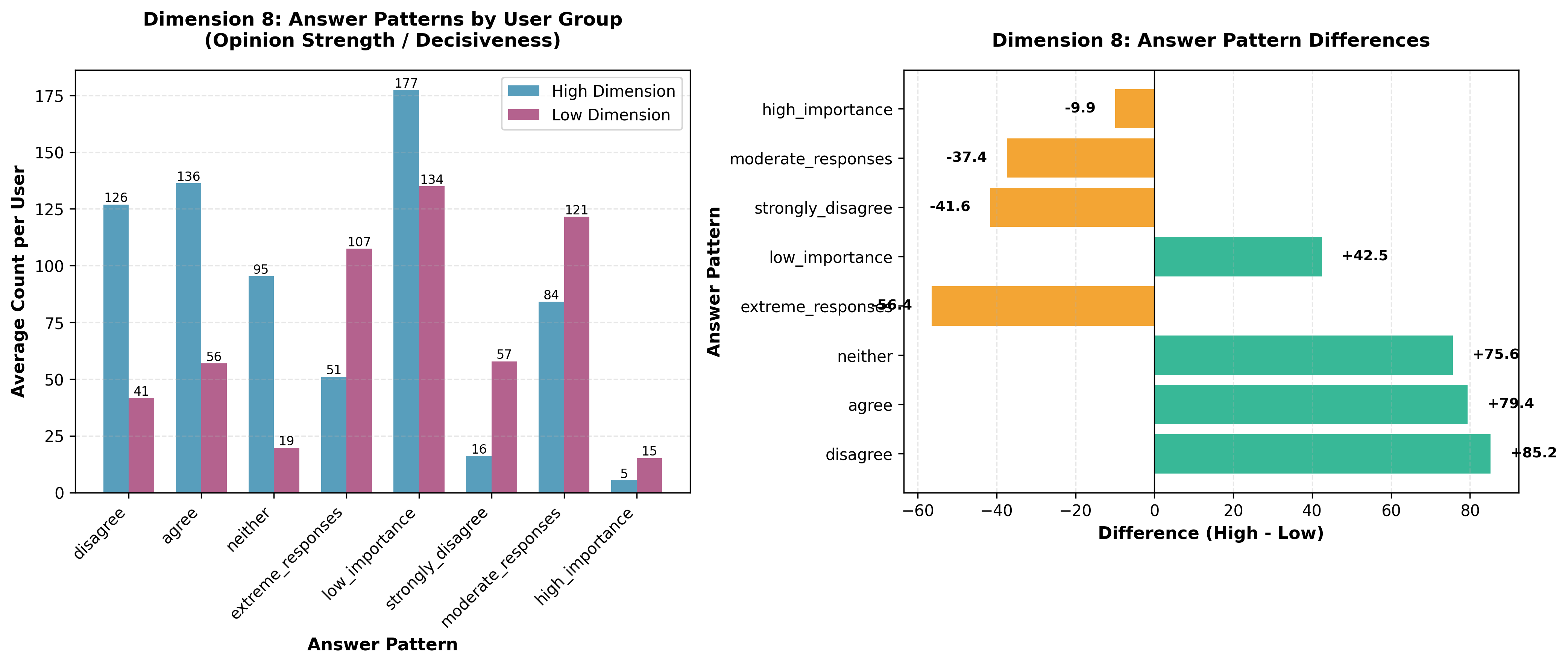}
  \caption{
  Response pattern comparison for users at the high and low extremes of latent Dimension~8.
  }
  \label{fig:appendix_dim8_patterns}
\end{figure}

\begin{figure}[t]
  \centering
  \includegraphics[width=\columnwidth]{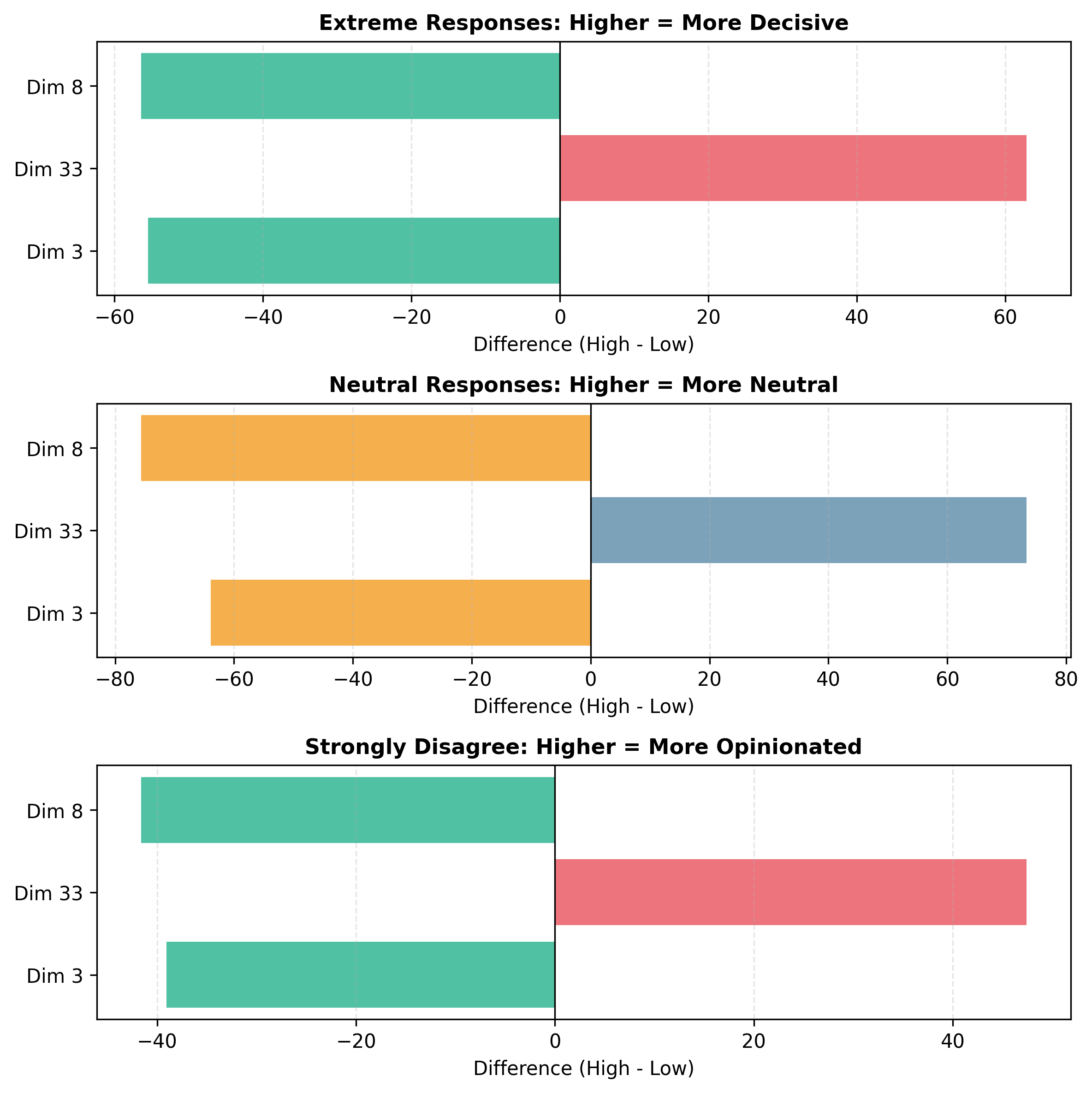}
  \caption{
  Summary comparison of response pattern differences across selected latent dimensions.
  Differences are computed between users at high and low extremes of each dimension.
  }
  \label{fig:appendix_dimension_summary}
\end{figure}

\end{document}